\documentclass{article}

\usepackage{arxiv}
\usepackage[utf8]{inputenc} 
\usepackage[T1]{fontenc}    
\usepackage{url}            
\usepackage{booktabs}       
\usepackage{amsfonts}       
\usepackage{nicefrac}       
\usepackage{microtype}      
\usepackage{lipsum}
\usepackage{subcaption}
\usepackage{graphicx}
\graphicspath{ {./images/} }
\usepackage{graphicx}
\usepackage{amsmath}
\usepackage{amssymb}
\usepackage{booktabs}
\usepackage{float}
\usepackage{multirow}

\usepackage[linesnumbered,ruled,vlined]{algorithm2e}
\usepackage{algpseudocode}

\usepackage[pagebackref,breaklinks,colorlinks]{hyperref}
\usepackage[capitalize]{cleveref}

\usepackage{cite}
\usepackage{textcomp}
\usepackage{xcolor}

\title{Identity-Focused Inference and Extraction Attacks on Diffusion Models}

\author{
Jayneel Vora$^{1}$, Aditya Krishnan$^{2}$, Nader Bouacida$^{1}$, Prabhu RV Shankar$^{3}$,  Prasant Mohapatra$^{4}$\\
$^{1}$Department of Computer Science, University of California, Davis\\
$^{2}$Department of Electrical and Computer Engineering, University of California, Davis\\
$^{3}$Department of Public Health Sciences, University of California Davis Health\\
$^{4}$Department of Computer Science, University of South Florida,Tampa\\
\texttt{$^{1}$\{jrvora,nbouacida\}@ucdavis.edu}\\
\texttt{$^{2}$adikrishnan@ucdavis.edu}\\
\texttt{$^{3}$rvpshankar@ucdavis.edu}\\
\texttt{$^{4}$pmohapatra@usf.edu}
}

\begin{document}
\maketitle

\begin{abstract} 
The increasing reliance on diffusion models for generating synthetic images has amplified concerns about the unauthorized use of personal data, particularly facial images, in model training. 
In this paper, we introduce a novel identity inference framework to hold model owners accountable for including individuals' identities in their training data. 
Our approach moves beyond traditional membership inference attacks by focusing on identity-level inference, providing a new perspective on data privacy violations. 
Through comprehensive evaluations on two facial image datasets, Labeled Faces in the Wild (LFW) and CelebA, our experiments demonstrate that the proposed membership inference attack surpasses baseline methods, achieving an attack success rate of up to 89\% and an AUC-ROC of 0.91, while the identity inference attack attains 92\% on LDM models trained on LFW, and the data extraction attack achieves 91.6\% accuracy on DDPMs, validating the effectiveness of our approach across diffusion models.

\end{abstract}


\section{Introduction}

Recent advancements in diffusion models~\cite{ho2020denoising,song2020denoising} have significantly transformed generative modeling, enabling breakthroughs in image~\cite{ho2022cascaded}, audio~\cite{liu2024audioldm}, and video~\cite{ho2022video} synthesis. 
These models have been widely adopted across industries such as healthcare~\cite{wolleb2022diffusion} and the creative arts~\cite{saharia2022palette} due to their ability to generate high-fidelity synthetic content. 

However, with the access to personal images from social media and other online data stores, concerns regarding the inclusion of sensitive data, particularly facial images~\cite{kim2023dcface}~\cite{huang2023collaborative}, without the knowledge or consent of the data owners have become increasingly prevalent. 
This issue raises significant challenges related to privacy, intellectual property, and the ethical use of personal data in AI systems.
A central challenge in this context is determining whether data related to a specific individual's identity was used to train these models. 

In this paper, we introduce the concept of \textit{identity inference}, which holds model owners accountable for the potential unauthorized use of personal data.
Unlike traditional membership inference, which seeks to determine whether a particular data point was part of the training set, identity inference focuses on detecting whether any known or unknown data point related to the individual's identity was used. 

\begin{figure*}[ht]
    \centering
    \includegraphics[width=\textwidth]{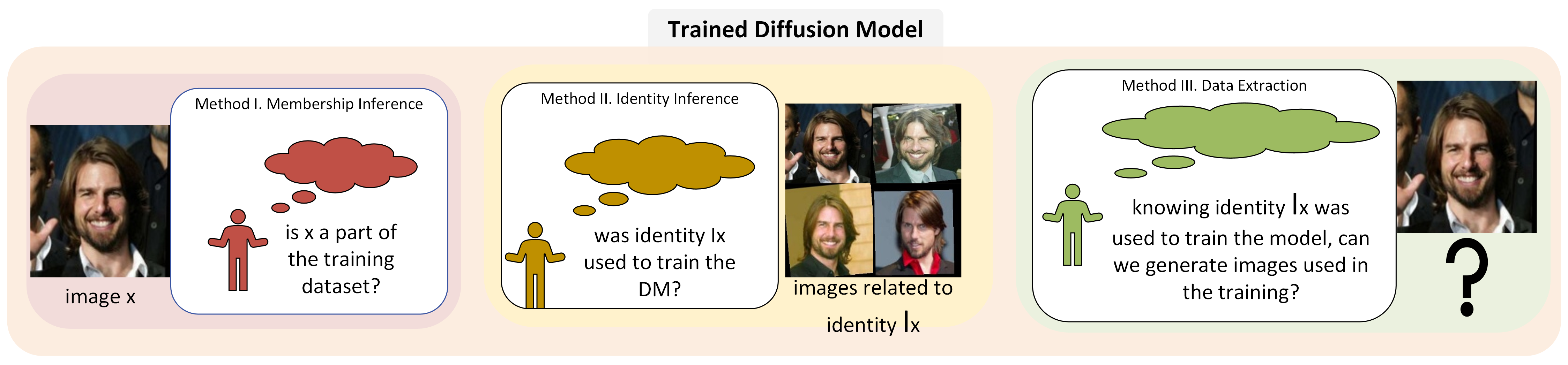}
    \caption{Overview of the denoising process in diffusion models with attack objectives: (1) Membership inference attack: determining if a specific query image was used in the training dataset. (2) Identity inference attack: verifying whether any data point related to the query image's identity was part of the training set. (3) Inferred identity generation attack: generating new data points related to the query image's identity used in the training dataset.}
    \label{fig:threat}
\end{figure*}

As diffusion models become more prominent, especially in domains involving sensitive data like facial images, the risk of training on unauthorized data becomes a growing concern~\cite{miernicki2021artificial}. Publicly available images online, for example, may be used without permission to train models capable of producing highly realistic synthetic images, potentially leading to significant privacy violations~\cite{bommasani2021opportunities}. It is thus essential to develop mechanisms that detect such unauthorized data use and ensure accountability, aligning with data privacy regulations.

Data owners can thus leverage identity inference to determine if their data is being used without authorization, without knowing the exact data~\cite{dealcala2024my} or anything about the dataset. Meanwhile, model owners can use it to prove or disprove any claims made by data owners.

In this paper, we systematically explore \textit{identity inference} within diffusion models trained on facial images, proposing a framework for assessing whether a specific identity was part of the training data. 
We emphasize that this framework is not an adversarial attack but a means to promote transparency and encourage ethical AI development. 
Our approach equips individuals with tools to verify if their data has been used without consent, pushing for more responsible data management practices by model developers by evolving privacy regulations.
\\

\textbf{Research Question:} \textit{Can we infer and(or) generate data related to a particular identity that was used in the training of a diffusion model?}
\\

To address this question, we define "data related to a particular identity" as any data associated with or linked to an identity in the training set without requiring exact matches to specific data points. 
We investigate this by exploring multiple attack objectives varying from inferring membership of particular data points to inferring the identity for a given list of query images and generating other potentially existing images in the training set.

Our proposed tools leverage the memorization tendencies of diffusion models to perform the inferences and extraction using single or multiple query images. 
Once an identity is inferred, we also demonstrate the generation of new data related to that identity. 

\subsection{Contributions}
The key contributions of this paper can be summarized as:

\begin{enumerate} 

\item \textbf{Membership Inference using Occluded Facial Regions}: We propose a novel membership inference attack designed specifically for diffusion models trained on facial images. 
This method leverages occluded facial regions to improve inference by focusing on how missing or masked features affect diffusion model sampling. 

\item \textbf{Formalization of Identity Inference}: We introduce a novel framework for identity inference in diffusion models, focusing on detecting whether a specific identity was part of the training data.
 This framework is grounded in a statistical analysis of reconstruction patterns and loss variability over multiple timesteps, enabling the detection of whether a specific identity was used during model training.

\item \textbf{Methods for Identity Inference and Image Extraction}: We develop membership and identity inference methods and data extraction methods to extract images related to identity in the training dataset, enabling individuals to verify whether their identity was used in a model's training set.

\item \textbf{Comprehensive Evaluation on Facial Image Datasets}: 

We extensively evaluate our methods using various popular facial image datasets, including Labeled Faces in the Wild (LFW) and CelebA. 
We evaluate the performance of our membership and identity inference using metrics such as accuracy, precision, recall, and AUC-ROC. The extraction method evaluations are reported utilizing the attack success rate.
Our results demonstrate the feasibility and effectiveness of our methods in different dataset scenarios.

\item \textbf{In-Depth Analysis of Key Influencing Factors}: 
We perform a detailed analysis of the factors influencing the success of membership and identity inference. 
These factors include the number of query images available, types of facial occlusions, diffusion model architecture, and number of denoising timesteps. 
We also explore the limitations of our approach and provide guidelines for future research on improving privacy-preserving diffusion models.
\end{enumerate}

\section{Related Work}
As diffusion models become increasingly popular for tasks such as image generation and text-to-image synthesis, they have also introduced new privacy challenges, particularly the risk of membership inference attacks (MIAs)~\cite{shokri2017membership}, which can expose whether specific data points were part of a model’s training set.

Early approaches primarily focused on loss-based metrics. The Naive Loss Method ~\cite{matsumoto2023membership} uses the training loss of the diffusion models to distinguish members from non-members by identifying timesteps with the most significant loss divergence. Similarly, PIA ~\cite{kong2023efficient} contrasts loss values between member and holdout holdoutto infer membership effectively. SecMI ~\cite{duan2023diffusion} introduces \emph{t-error}, measuring posterior estimation errors at each timestep under the assumption that members exhibit smaller t-errors compared to non-members.

Generative prior-based methods offer alternative strategies. The Degrade-Restore-Compare (DRC) method ~\cite{fu2024model} partially degrades input images, restores them via reverse diffusion, and compares the originals with restorations using semantic embeddings to assess membership. ReDiffuse ~\cite{li2024towards} generates multiple noisy versions of an input image, averages the results, and determines membership based on similarity. Additionally, Li et al.~\cite{li2024unveiling} utilize structural similarity indices between original and corrupted images for inference.

In text-to-image models, CLiD ~\cite{zhai2024membership} leverages conditional overfitting by measuring divergence in likelihood estimations under various textual prompts to perform MIAs. Gradient-based attacks have also been explored, such as the Gradient Subspace Aggregation (GSA) ~\cite{pang2023white}, aggregating gradient information across timesteps and layers to distinguish members from non-members.

Black-box attacks offer another model. Pang et al. ~\cite{pang2023black} compute similarities between query and generated images using feature extraction and statistical aggregation without accessing internal model parameters. Other methods exploit discrepancies between generated and training data, as seen in Wu et al. ~\cite{wu2022membership}, who utilize pixel-level and semantic fidelity differences, and Tang et al. ~\cite{tang2023membership}, who apply quantile regression to predict reconstruction loss distributions for membership determination.

Likelihood-based techniques, such as the Likelihood Ratio Attack (LiRA) ~\cite{carlini2023extracting}, train shadow models to derive confidence value distributions, enabling membership inference without high computational costs due to multiple model training.

Despite the extensive research on MIAs, most studies concentrate on determining the presence of specific data points within the training set. These methods vary in their assumptions, computational requirements, and effectiveness, highlighting the inherent vulnerability of diffusion models to privacy breaches.

Our work extends beyond membership inference attacks by determining whether data related to a given identity was used during training rather than focusing on individual data points. 
Our approach highlights a broader privacy risk in generative models, targeting the presence of identities within the training data. Our proposed framework fills a significant gap in the privacy evaluation of diffusion models and emphasizes an exciting direction for holding diffusion model owners accountable for the data used.

\section{Background}

Diffusion models, inspired by non-equilibrium thermodynamics\cite{sohl2015deep}, offer a robust generative framework by iteratively transforming data through a noise-adding process, unlike the direct encoding-decoding approaches of Variational Autoencoders(VAEs)~\cite{kingma2013auto} and the adversarial training dynamics of Generative Adversarial Networks (GANs)~\cite{goodfellow2020generative}.

In Denoising Diffusion Probabilistic Models (DDPMs)~\cite{ho2020denoising}, the forward process progressively corrupts a data sample \(x_0\) by adding Gaussian noise over \(T\) timesteps, governed by a variance schedule \(\{\beta_t\}_{t=1}^T\), where each \(\beta_t \in (0, 1)\). This process generates noisy samples \(x_1, x_2, \dots, x_T\), with \(x_T\) approaching an isotropic Gaussian distribution. The transition at each step \(t\) is modeled as:
\begin{equation}
q(x_t \mid x_{t-1}) = \mathcal{N}\left( x_t; \sqrt{1 - \beta_t} \, x_{t-1}, \beta_t \mathbf{I} \right),
\label{eq:forward}
\end{equation}
where \(\alpha_t = 1 - \beta_t\) controls the variance of the noise added at each timestep, and the cumulative product \(\bar{\alpha}_t = \prod_{s=1}^t \alpha_s\) reflects the scaling factor applied to the original data \(x_0\) at timestep \(t\), determining how much of the original signal contributes to \(x_t\). As \(t\) increases, \(\bar{\alpha}_t\) decreases, resulting in a gradual transition of the data towards Gaussian noise~\cite{ho2020denoising}.

The reverse process aims to recover the original data \(x_0\) from \(x_T\) through an iterative denoising process. This is parameterized by learned conditional distributions:
\begin{equation}
p_\theta(x_{t-1} \mid x_t) = \mathcal{N}\left( x_{t-1}; \mu_\theta(x_t, t), \hat{\Sigma}_\theta(x_t, t) \right),
\label{eq:reverse}
\end{equation}
where \(\mu_\theta\) is the predicted mean, and \(\hat{\Sigma}_\theta\) represents the variance. Typically, \(\hat{\Sigma}_\theta\) can either be learned or fixed; we fix it. The training objective is to minimize the difference between the noise added during the forward process and the predicted noise \(\epsilon_\theta\):
\begin{equation}
L(\theta) = \mathbb{E}_{x_0, t, \epsilon} \left[ \left\| \epsilon - \epsilon_\theta\left( x_t, t \right) \right\|_2^2 \right],
\end{equation}
where \(x_t = \sqrt{\bar{\alpha}_t} \, x_0 + \sqrt{1 - \bar{\alpha}_t} \, \epsilon\) and \(\epsilon \sim \mathcal{N}(0, \mathbf{I})\).

During generation, the reverse process begins with \(x_T \sim \mathcal{N}(0, \mathbf{I})\), and each step \(x_{t-1}\) is updated as follows:
\begin{equation}
x_{t-1} = \frac{1}{\sqrt{\alpha_t}} \left( x_t - \frac{\hat{\beta}_t}{\sqrt{1 - \bar{\alpha}_t}} \, \epsilon_\theta(x_t, t) \right) + \hat{\beta}_t z_t,
\end{equation}
where \(\hat{\beta}_t = \frac{1 - \bar{\alpha}_t}{1 - \bar{\alpha}_{t-1}} \beta_t\) controls the variance, and \(z_t \sim \mathcal{N}(0, \mathbf{I})\).

\textbf{Denoising Diffusion Implicit Models (DDIMs)}~\cite{song2020denoising} offer a variant of this sampling process, providing a deterministic mapping from \(x_T\) to \(x_0\) while eliminating stochasticity by directly parameterizing the reverse process.

\textbf{Latent Diffusion Models (LDMs)}~\cite{rombach2022high} further enhance the efficiency of diffusion models by operating in a lower-dimensional latent space, learned through an autoencoder. 
This transformation into a compressed latent space reduces the required timesteps, enabling faster synthesis while preserving high-quality outputs. 
By operating in this reduced space, LDMs reduce unnecessary noise propagation and improve computational efficiency.

\section{Problem Framework}
In this section, we formalize three attacks on generative diffusion models: \emph{membership inference}, \emph{identity inference}, and \emph{data extraction}. 
These attacks provide a mechanism for auditors and data owners to assess claims about the presence or absence of specific data identities or data points in a diffusion model's training dataset.

Consider a dataset \(\mathcal{D} = \{(x_i, y_i)\}_{i=1}^{D}\), where each image \(x_i \in \mathbb{R}^{H \times W \times C}\) represents an image tensor with height \(H\), width \(W\), and \(C\) color channels (typically \(C=3\) for RGB images). 
Each image is labeled with an identity \(y_i \in \mathcal{I} = \{I_1, I_2, \dots, I_N\}\). 
The dataset is randomly partitioned into two disjoint subsets: a training set \(\mathcal{D}_{\text{train}}\) and a holdout  \(\mathcal{D}_{\text{hold}}\) with samples drawn independently and uniformly. \(\mathcal{D}_{\text{train}}\) and \(\mathcal{D}_{\text{hold}}\) are such that,  \(\mathcal{D} = \mathcal{D}_{\text{train}} \cup \mathcal{D}_{\text{hold}}\) and \(\mathcal{D}_{\text{train}} \cap \mathcal{D}_{\text{hold}} = \emptyset\). 
Partitioning is performed at the image level, allowing images from the same identity to appear in both subsets.

We train a diffusion model \(M_\theta\) on \(\mathcal{D}_{\text{train}}\), aiming to approximate the data distribution \( p_\text{data}(x) \)..
The core hypothesis underlying our attacks is based on the understanding from previous studies\textbf{CITE} that trained diffusion models \(M_\theta\) reconstruct images similar to training images more consistently than non-training images, thus enabling inference attacks. 

\textbf{Assumptions:}
We assume the following about the adversary's knowledge:
\begin{enumerate}
    \item The adversary has full white-box access to the sampling process of the model \(M_\theta\), including its architecture, weights, and loss functions.
    \item The adversary has access to a query image \(x_q\) omodel'st of query images \(\mathcal{X}_q\) but no direct access to the training dataset \(\mathcal{D}_{\text{train}}\).
    \item The adversary can perform multiple queries to the model, observing its behavior, but does not know whether a query image(s) belongs to the training dataset.
\end{enumerate}

Based on these assumptions, we define the following attack objectives:

\textbf{Membership Inference Attack(MIA):}
In the membership inference attack, the adversary aims to determine whether a specific query image \(x_q\) was used in training the diffusion model \(M_\theta\). The adversary queries our framework with \(x_q\), which analyzes the model's sampling process to infer whether \(x_q \in \mathcal{D}_{\text{train}}\). 
In the identity inference attack, the adversary seeks to determine whether an identity \(I_q\), associated with a set of images, was used during training. 
Given a set of query images \(\mathcal{X}_q = \{ x_q^1, x_q^2, \dots, x_q^M \}\), where all images are linked to the same identity \(I_q\), the adversary infers the presence of \(I_q\) in the training dataset \(\mathcal{D}_{\text{train}}\). 
This attack is more potent than membership inference when multiple images from the same identity are available.

\textbf{Data Extraction Attack:}
In the data extraction attack, the adversary aims to extract additional data points related to a known identity \(I_q\) used in the training dataset. 
Given access to a query image \(x_q\), the adversary attempts to generate images that are similar to the images related to the identity \(I_q\) in \(\mathcal{D}_{\text{train}}\).

\section{Methodology}
In this section, we describe the methodology for inferring membership and identity in diffusion models, specifically through the \emph{Membership Inference Attack (MIA)} and the \emph{Identity Inference Attack (IIA)}. These attacks rely on the intuition that models behave differently on training data than unseen data, allowing the adversary to model differences about data presence in the training set. We also describe the \emph{Inferred Identity Data Extraction Attack}, where the adversary attempts to generate images representative of an identity inferred from the model.

\subsection{Creating Occlusion Masks}
We use occlusion masks for every attack we propose for the eyes, nose, and mouth. 
Given a query image $x_q \in \mathbb{R}^{H \times W \times C}$, we apply a carefully designed set of facial masks $\{M_i\}_{i=1}^N$. 
Each mask occludes specific facial regions, such as the eyes, nose, or mouth, allowing us to probe the model's reconstruction behavior on different facial features. 
To create an exhaustive yet computationally manageable set of masks, we use a variant of Reference Heatmap Transformer(RHT)~\cite{wan2023precise}, a facial landmark detection to identify vital facial regions—eyes, nose, mouth, forehead, cheeks, and chin. 
We then generate masks by occluding individual regions, grouping regions, and randomly selecting facial patches to simulate various occlusions. This strategy ensures coverage of critical facial areas while keeping the number of masks $N$ practical for computational efficiency.

The masked images are represented as:
\begin{equation}
    x_q^{M_i} = x_q \odot M_i,
    \label{eq:masks}
\end{equation}
where $ \odot$ denotes element-wise multiplication and $ M_i$ is a binary mask. 
The facial masks are crucial for isolating region-specific behaviors in the reconstruction process.

\subsection{Membership Inference}

We propose a novel membership inference technique for trained diffusion models leveraging the variability in reconstruction losses during the model's iterative sampling process. 
We compute a continuous confidence score $C(x_q) \in [0,1]$  indicating the likelihood that a given query image was part of the training set without requiring access to the original training data or auxiliary datasets.

By analyzing how the model reconstructs different masked regions, we can infer the model's familiarity with the input image. 
We hypothesize that a model trained on a particular face is expected to produce more consistent reconstructions across various masked versions of that face, resulting in lower variability in the model construction losses.

To account for varying amounts of visible pixels due to masking, we normalize the reconstruction losses by the number of unmasked pixels in each $M_i$. Specifically, we define a normalization factor $N_i = \sum_{p} M_i(p)$, where $p$ indexes over all pixels. 
This normalization ensures that the reconstruction losses are comparable across different masks.

For a query image, $x_q$, masks are applied using Eq.~\ref{eq:masks} to obtain masked images $x_q^{M_i}$. Next, the forward diffusion process is applied to obtain a noise version. Further, a sequence of reconstructions $\hat{x}_q^{M_i}(t)$ across timesteps $t$, refining the noisy input through the denoising process using Eq.\ref{eq:reverse}. The reconstruction loss at each timestep is computed for each denoised, masked image using Eq.~\ref{eq:mia_recon}
\begin{equation}
    E_i(t) = \frac{1}{N_i} \left| x_q^{M_i} - \hat{x}_q^{M_i}(t) \right|_2,
    \label{eq:mia_recon}
\end{equation}
where the normalization by $N_i$ adjusts for the varying amounts of visible pixels.

To capture the model's behavior throughout the sampling process, we analyze the sequence of losses $\{E_i(t)\}_{t=1}^T$ using statistical measures. Specifically, we compute the coefficient of variation (CV), skewness ($S_i$), and the mean rate of change ($\Delta E_i$) for each masked input:
\begin{align}
    \text{CV}_i &= \frac{\sigma_{E_i}}{\mu_{E_i}}, \\
    S_i &= \frac{1}{T} \sum_{t=1}^{T} \left( \frac{E_i(t) - \mu_{E_i}}{\sigma_{E_i}} \right)^3, \\
    \Delta E_i &= \frac{1}{T-1} \sum_{t=2}^{T} \left| E_i(t) - E_i(t-1) \right|,
\end{align}
where $\mu_{E_i}$ and $\sigma_{E_i}$ are the mean and standard deviation of $E_i(t)$ over timesteps.

We aggregate these statistical measures across all masks to compute the confidence score $C(x_0)$, defined as the confidence score in Eq.~\ref{eq:confidence_score}.
\begin{equation}
    C(x_q) = \frac{1}{1 + \frac{1}{N} \sum_{i=1}^{N} \left( \text{CV}_i + |S_i| + \Delta E_i \right)}.
    \label{eq:confidence_score}
\end{equation}
This formulation ensures that $C(x_0)$ is a continuous value between 0 and 1, inversely related to the average variability and asymmetry in the reconstruction losses. A lower variability and more consistent reconstruction—indicative of a training set member—will result in a higher confidence score. In contrast, higher variability suggests the image is less familiar to the model.


\subsection{Identity Inference}

We now introduce the identity inference attack building on the frameworks set by the bymodel's
Unlike MIAs, where individual images are analyzed to determine membership, this attack operates at the identity level, inferring whether images of a specific identity $\mathcal{I}$ were part of the training set. 
The attack works by analyzing the reconstruction behavior of the model across multiple query images, all associated with the same identity $\mathcal{I}$. 
Specifically, a set of query images, $\{ x_0^{(k)} \}_{k=1}^K$, each representing the same identity $\mathcal{I}$, is subjected to random occlusions and then passed through the model. 
The reconstruction errors are analyzed over multiple timesteps to capture the temporal variability in the model's behavior for these query images. 
By examining how consistent the model is in reconstructing images of the same identity over different noise levels, we compute a score that quantifies the likelihood that images of this identity were included in the training dataset.

Let $\mathcal{I}$ represent a specific identity, such as a person’s face, and let $\{ x_0^{(k)} \}_{k=1}^K$ be the set of query images of identity $\mathcal{I}$, where each image $x_0^{(k)} \in \mathbb{R}^{H \times W \times C}$ is reconstructed across multiple timesteps. The objective is to estimate whether images of this identity were present in the model’s training set. We denote the reconstruction at timestep $t$ for each query image as $\hat{x}_t^{(k)}$, and the corresponding reconstruction error is defined as the Euclidean distance between the original and reconstructed image at timestep $t$, as follows:

\begin{equation}
    E_t^{(k)} = \left\| x_0^{(k)} - \hat{x}_t^{(k)} \right\|_2, \quad t = 1, 2, \ldots, T,
\end{equation}
\label{eq:reconstruction_error}
where $k$ indexes the query images and $t$ indexes the timesteps. This gives us a sequence of reconstruction errors $\{ E_t^{(k)} \}_{t=1}^T$ for each query image, representing the temporal evolution of the reconstruction errors as the reverse diffusion process progresses.

We compute the mean and variance of the reconstruction errors across the timesteps for each query image. The mean reconstruction error for the $k$-th query image is computed as:

\begin{equation}
    \mu_E^{(k)} = \frac{1}{T} \sum_{t=1}^{T} E_t^{(k)},
\label{eq:mean_reconstruction_error}
\end{equation}
and the variance in reconstruction errors for the same query image is given by:

\begin{equation}
    \sigma_E^{(k)} = \sqrt{ \frac{1}{T} \sum_{t=1}^{T} \left( E_t^{(k)} - \mu_E^{(k)} \right)^2 }.
\label{eq:variance_reconstruction_error}
\end{equation}

We extend this analysis to the entire set of query images, calculating both the overall mean reconstruction error and the overall variance in reconstruction errors as:

\begin{equation}
    \mu_E = \frac{1}{K} \sum_{k=1}^{K} \mu_E^{(k)}, \quad
    \sigma_E = \frac{1}{K} \sum_{k=1}^{K} \sigma_E^{(k)}.
\label{eq:overall_mean_and_variance}
\end{equation}

The next step is to compute a score that quantifies how consistent the model is in reconstructing the set of query images over time. We propose the **Score for Identity Inference (S\textsubscript{II})**, denoted as $S_{\text{II}}$, which is computed as follows:

\begin{equation}
    S_{\text{II}} = \exp\left( - (\sigma_E + \mu_E) \right).
\label{eq:sii}
\end{equation}

This score considers both the variability (through $\sigma_E$) and the average quality (through $\mu_E$) of the model's reconstructions.
The score $S_{\text{II}}$ is designed to lie within the range $[0, 1]$, where value closer to $1$ indicates perfect reconstruction consistency and minimal error, strongly suggesting that identity $\mathcal{I}$ was part of the training set. A value of $0$ indicates high reconstruction variability or error, implying that the identity was not part of the training data. The exponential form of the score ensures that small increases in the reconstruction error or variability have an exponentially decreasing impact on the score, effectively penalizing both high error and high variability. 

Mathematically, the score is bounded by $0 \leq S_{\text{II}} \leq 1$. When $\sigma_E + \mu_E = 0$, the score attains its maximum value of $S_{\text{II}} = 1$, corresponding to perfect consistency and accuracy, which implies that identity $\mathcal{I}$ was almost certainly included in the training data. As $\sigma_E + \mu_E$ increases, the score decreases exponentially, approaching $0$ in the limit as the reconstruction errors become large or highly variable, indicating that identity $\mathcal{I}$ was likely not in the training set.

If $S_{\text{II}} \geq 0.5$, we infer that identity $\mathcal{I}$ was likely part of the training set. A score closer to 1 suggests that the model does not consistently reconstruct this identity's images; hence, the identity was likely not part of the training data. This threshold balances the need for high confidence in the inference while allowing for some variability in reconstruction errors.

\subsection{Data Extraction Attack}

The goal of this attack is to generate images from the diffusion model's training dataset that are related to a specific identity, given a query image. 
We introduce a novel approach that leverages facial masks and multi-seed generation, followed by clustering to identify the most likely memorized images corresponding to the given identity.

We begin by applying feature-preserving facial masks to the query image \( x_q \). These masks preserve specific facial regions, such as the eyes, nose, and mouth, while occluding all other parts of the face. 
The regions to be preserved are randomly selected for each mask, ensuring diverse masked representations of the same query image. 
For each masked version \( x_q^{M_i} \), we apply the forward diffusion process to obtain a noise version. Once the noisy images are generated, we apply the reverse diffusion process using multiple random seeds \( \{ s_j \}_{j=1}^{S} \). Each seed helps in generating a varied reconstruction for the same masked query image, resulting in multiple diverse outputs:
\begin{equation}
\hat{x}_q^{M_i}(s_j), \quad j = 1, \dots, S.
\end{equation}

Next, for each set of reconstructions \( \hat{x}_0^{M_i}(s_j) \) generated from a given mask, we analyze the consistency of the reconstructions across different seeds. Specifically, we focus on the facial regions preserved by the masks. By identifying the commonly occurring patterns or features of identity in these regions, we can infer which reconstructions are most likely to correspond to images in the training dataset.

To enhance the robustness of this identification, we generate a large set of images, typically \( 100 \) reconstructions, across multiple masks and seeds. Then, we use K-means clustering algorithm to group the generated images based on their similarity in the masked regions. The clustering is performed as follows:

Let the set of all generated reconstructions be denoted as \( \mathcal{X} = \{ \hat{x}_0^{M_i}(s_j) \} \), and let the corresponding masked regions for each image be represented as feature vectors \( f(\hat{x}_0^{M_i}(s_j)) \). These feature vectors are extracted based on pixel values in the preserved facial regions.

We define the objective function for k-means as:
\begin{equation}
\min_{\{\mu_k\}_{k=1}^{K}} \sum_{k=1}^{K} \sum_{\hat{x}_0 \in C_k} \| f(\hat{x}_0) - \mu_k \|^2,
\end{equation}
where \( K \) is the number of clusters, \( C_k \) represents the set of images in the \( k \)-th cluster, and \( \mu_k \) is the centroid of the cluster.

We initialize the centroids \( \{ \mu_k \} \) using k-means++ initialization\textbf{cite} and iteratively update the clusters by minimizing the distance between each image’s feature vector and the cluster centroid. The algorithm terminates when the assignments no longer change, and each cluster \( C_k \) represents a distinct image candidate related to the identity from the model's training set.

The representative image of each cluster is chosen as the cluster centroid \( \mu_k \), which minimizes the sum of squared distances within the cluster. In our work, we use 10 clusters after sampling 100 images after empirically testing other combinations.

\section{Evaluations}

In this section, we evaluate the performance of our attacks across various datasets, models, and settings.
\subsection{Evaluation Setup}
\subsubsection{Datasets}
Our evaluation experiments were conducted on three well-known facial image datasets: 
\begin{enumerate}
    \item  LFW dataset ~\cite{huang2008labeled} consists of 13,233 face images, sourced from the internet, cropped to a resolution of 250x250 pixels. It contains 5,749 unique individuals, with 1,680 of those individuals having two or more images.     
     \item CelebA dataset ~\cite{liu2015deep} comprises 202,599 cropped and aligned facial images featuring 10,177 different identities.

\end{enumerate}
Each dataset was carefully partitioned, with 60\% of the data used for training and the remaining 40\% set aside as a holdout sholdout the purpose of our experiments, the datasets were split into training sets, denoted as $\mathcal{D}_{\text{train}}$image'soldout sets, denoted as $\mathcal{D}_{\text{hold}}$, ensuring that the identities were mutually exclusive between the two sets. 
For CelebA, 8,000 identities were assigned to the training set, and 2,177 images were reserved for the holdout. Similarly, LFW was partitioned with 4,500 identities in the training set and 1,249 in the holdout set. The disjoint identity partitioning ensures that there is no overlap between the individuals present in the training and holdout sets.
\subsubsection{Model Training}
For each dataset, an unconditional diffusion model denoted as $M_\theta$, was trained using the respective training set $\mathcal{D}_{\text{train}}$. Specifically, we employed the DDPM ~\cite{ho2020denoising}, DDIM, and LDM models, following the same settings outlined in their original papers. All models were trained on a server with a 48GB Nvidia GPU to ensure sufficient computational resources for effective training.
\subsubsection{Threat Model Assumptions}
Regarding the threat model, we assume an adversary with significant capabilities. 
The adversary is granted white-box access to the trained model $M_\theta$, including complete visibility of its parameters and architecture. 
Additionally, the adversary is assumed to have detailed knowledge of the diffusion process and noise schedule $\beta_t$. 
They also have access to auxiliary data, which includes query images $x_q$ that may or may not be present in the training dataset $\mathcal{D}_{\text{train}}$ for membership inference, as well as images of identities $I_q$ for identity inference, where the goal is to determine whether these identities were included in $\mathcal{D}_{\text{train}}$.

\begin{table*}[]
\resizebox{0.9\textwidth}{!}{%
\begin{tabular}{cccccccccc}
\hline
\multirow{2}{*}{\textbf{Diffusion Model}} &
  \multirow{2}{*}{\textbf{Approach}} &
  \multicolumn{4}{c|}{\textbf{CelebA}} &
  \multicolumn{4}{c}{\textbf{LFW}} \\ \cline{3-10} 
 &
   &
  \textbf{Accuracy} &
  \textbf{Precision} &
  \textbf{Recall} &
  \multicolumn{1}{c|}{\textbf{AUC}} &
  \textbf{Accuracy} &
  \textbf{Precision} &
  \textbf{Recall} &
  \textbf{AUC} \\ \hline
\multirow{3}{*}{DDPM} & SecMI & 0.61 & 0.60 & 0.62 & \multicolumn{1}{c|}{0.63} & 0.79 & 0.78 & 0.80 & 0.82 \\ \cline{2-10} 
                      & DRC   & 0.64 & 0.63 & 0.65 & \multicolumn{1}{c|}{0.66} & 0.82 & 0.81 & 0.83 & 0.84 \\ \cline{2-10} 
                      & \textbf{Our}   & \textbf{0.70} & \textbf{0.69} & \textbf{0.68} & \multicolumn{1}{c|}{\textbf{0.72}} & \textbf{0.89} & \textbf{0.88} & \textbf{0.90} & \textbf{0.91} \\ \hline
\multirow{3}{*}{DDIM} & SecMI & 0.63 & 0.62 & 0.64 & \multicolumn{1}{c|}{0.65} & 0.78 & 0.77 & 0.79 & 0.80 \\ \cline{2-10} 
                      & DRC   & 0.67 & 0.66 & \textbf{0.68} & \multicolumn{1}{c|}{0.69} & 0.85 & 0.84 & 0.86 & 0.87 \\ \cline{2-10} 
                      & \textbf{Our}   & \textbf{0.68} & \textbf{0.66} & 0.67 & \multicolumn{1}{c|}{\textbf{0.70}} & \textbf{0.87} & \textbf{0.86} & \textbf{0.88} & \textbf{0.89} \\ \hline
\multirow{3}{*}{LDM}  & SecMI & 0.60 & 0.59 & 0.61 & \multicolumn{1}{c|}{0.62} & 0.76 & 0.75 & 0.77 & 0.78 \\ \cline{2-10} 
                      & DRC   & 0.65 & 0.64 & \textbf{0.66} & \multicolumn{1}{c|}{0.67} & 0.81 & 0.80 & 0.82 & 0.83 \\ \cline{2-10} 
                      & \textbf{Our}   & \textbf{0.69} & \textbf{0.67} & 0.65 & \multicolumn{1}{c|}{\textbf{0.71}} & \textbf{0.88} & \textbf{0.87} & \textbf{0.89} & \textbf{0.90} \\ \hline
\end{tabular}%
}
\caption{
Performance comparison of membership inference attacks on the LFW and CelebA datasets, evaluated across metrics including accuracy, precision, recall, and AUC-ROC. Our method consistently outperforms the baseline approaches, particularly excelling on the LFW dataset.}
\label{tab:mia}
\end{table*}
\begin{figure}
    \centering
    \includegraphics[width=0.8\linewidth]{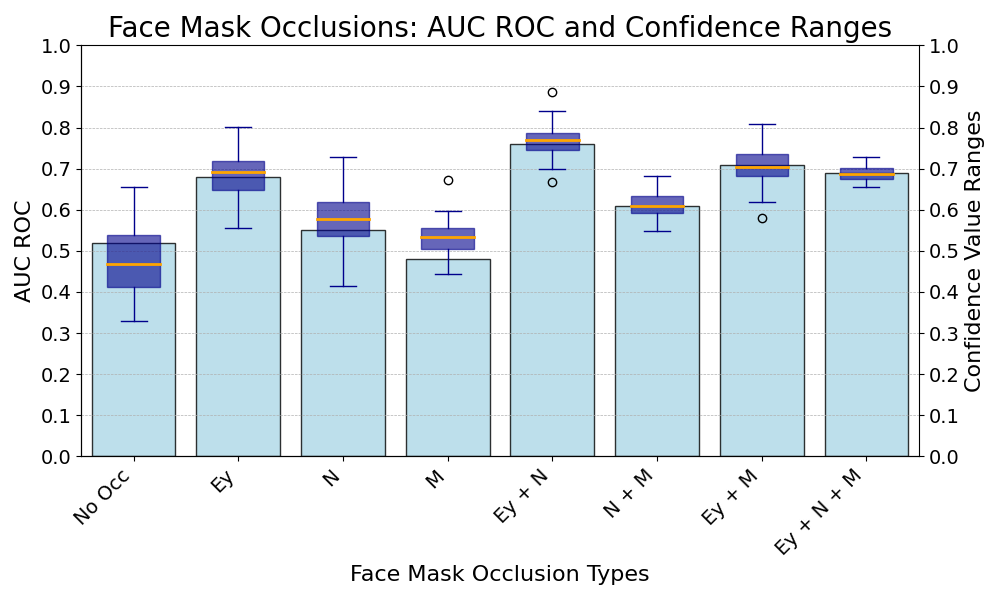}
    \caption{AUC-ROC and confidence value distributions for different face mask occlusion types on the CelebA dataset. The method shows the highest ASR for 'Ey' (eyes occluded), while occlusions involving the mouth (M) lead to reduced ASR. Confidence value ranges highlight the robustness of the inference mechanism for partial occlusions.}
    \label{fig:mia_masks}
\end{figure}
\subsection{Membership Inference}

To rigorously evaluate the performance of our proposed membership inference attack, we conducted experiments on diffusion models trained with facial image datasets (CelebA and LFW). We randomly sampled 100 images from each dataset and computed membership inference metrics-  accuracy, precision, recall, and AUC-ROC. The results are averaged over seven independent runs, and our method is compared against two established baselines: SecMI~\cite{duan2023diffusion} and DRC~\cite{fu2024model}, which are white-box attacks relying on reconstruction loss thresholds. The same query images were used for all methods to ensure consistent evaluation.

The results are detailed, and a threshold of 0.6 is used for the confidence value.in Table~\ref{tab:mia}, demonstrate that our method consistently outperforms the baseline approaches across all evaluated metrics. For instance, on the CelebA dataset, our method achieves an accuracy of 0.70, precision of 0.69, and AUC-ROC of 0.72, surpassing SecMI and DRC, which yield accuracies of 0.61 and 0.64, respectively. Similarly, on the LFW dataset, our approach attains an accuracy of 0.89 and an AUC-ROC of 0.91, significantly outperforming the baselines (SecMI achieves an AUC of 0.82 and DRC 0.84).

We computed the confidence score $C(x_q)$ for each query image using the confidence score defined in Eq.~\ref{eq:confidence_score}. This confidence-based approach provides a continuous measure of inference certainty, unlike the binary decisions employed by SecMI and DRC. Reconstruction losses were calculated using the same diffusion model $M_\theta$ for the baseline methods, with thresholds set according to their respective literature. By analyzing the distributions of confidence scores across occlusion types, we determined an appropriate threshold(0.6) for further analysis, ensuring that the selection process accounts for the AUC-ROC and the variability of confidence values. Identifying a threshold empirically allows us to make the data extraction attacks more potent in the following subsections.
\begin{figure*}[]
    \centering
    \begin{subfigure}[b]{0.48\textwidth}
        \centering
        \includegraphics[width=\textwidth]{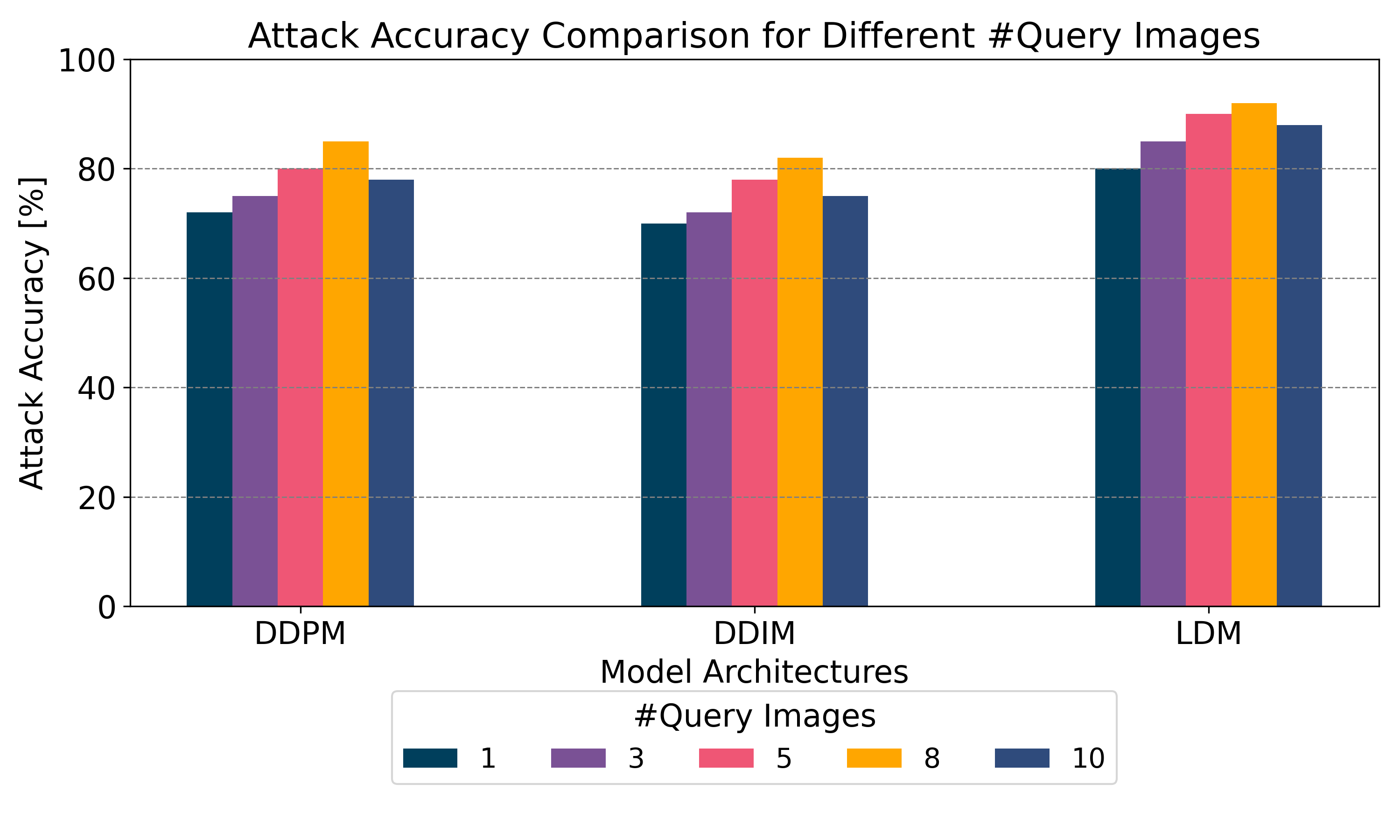}
        \caption{Attack Accuracy Comparison}
        \label{fig:asr}
    \end{subfigure}
    \begin{subfigure}[b]{0.48\textwidth}
        \centering
        \includegraphics[width=\textwidth]{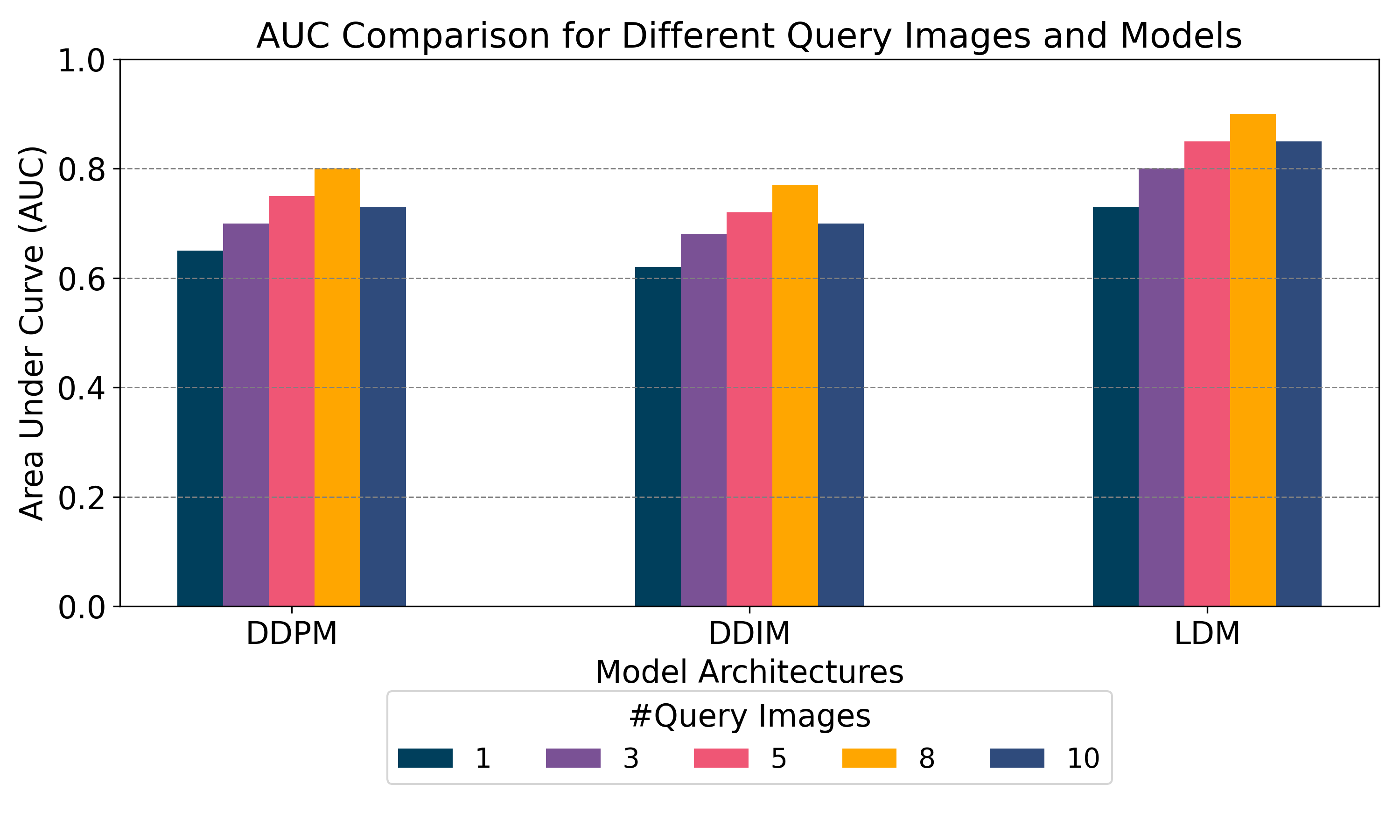}
        \caption{Area Under Curve (AUC) Comparison}
        \label{fig:auc}
    \end{subfigure}
    \caption{Comparison of Attack Accuracy and Area Under Curve (AUC-ROC) across different diffusion models (DDPM, DDIM, LDM) using various number of query images. The diffusion models are trained on the LFW. The figure illustrates the performance of identity inference attacks using 1, 3, 5, 8, and 10 query images.}
    \label{fig:comparison}
\end{figure*}
\subsection{Identity Inference}

\begin{figure*}[ht]
    \centering
    \begin{subfigure}[b]{0.32\textwidth}
        \centering
        \includegraphics[width=\textwidth]{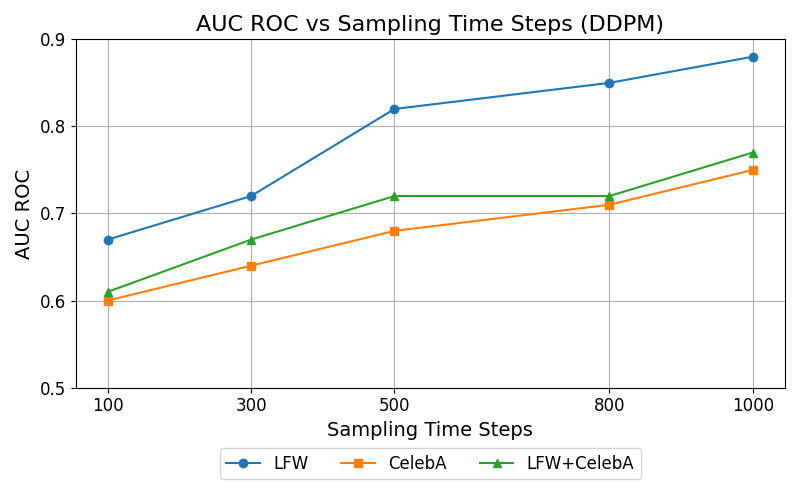}
        \caption{DDPM models, with 1000 sampling steps}
        \label{fig:plot1}
    \end{subfigure}
    \hfill
    \begin{subfigure}[b]{0.32\textwidth}
        \centering
        \includegraphics[width=\textwidth]{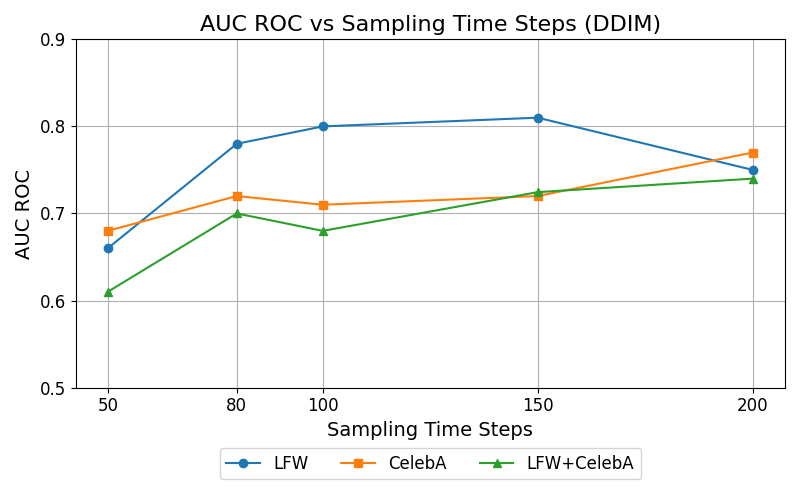}
        \caption{DDIMs models, with 200 sampling steps}
        \label{fig:plot2}
    \end{subfigure}
    \hfill
    \begin{subfigure}[b]{0.32\textwidth}
        \centering
        \includegraphics[width=\textwidth]{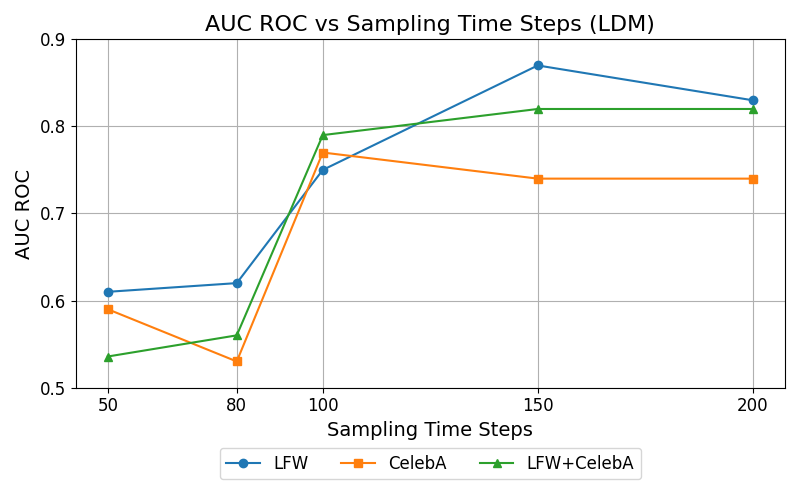}
        \caption{LDM model, with 200 sampling steps}
        \label{fig:plot3}
    \end{subfigure}

    \caption{AUC-ROC vs. Sampling Time Steps for Different Datasets: LFW, CelebA, and LFW+CelebA. For query image lists of size 5.}
    \label{fig:attack_success_rate}
\end{figure*}

In this section, we evaluate the effectiveness of our identity inference attack across different diffusion models: DDPM, DDIM, and LDM. 
Since this is the first identity inference attack of its kind, we do not have any prior baselines for comparison other than random guessing. 
The attack evaluates the ability to infer the membership of an identity in the training dataset, given a list of query images.

Figure~\ref{fig:asr} shows the attack accuracy, and Figure~\ref{fig:auc} shows the AUC-ROC as the number of query images increases from 1 to 10. 
The attack shows the highest effectiveness for LDMs, achieving the best AUC and accuracy across the range of query images. DDPM performs reasonably well, while DDIM performs slightly better than random guessing. 
Specifically, we observe that with just 5 query images, the attack on  LDM achieves an AUC-ROC close to 0.85, while on DDPM, it approaches 0.80. On the other hand, it consistently underperforms on DDIMs, hovering above the baseline of random guessing across all numbers of query images.

\begin{figure}
    \centering
    \includegraphics[width=0.7\columnwidth]{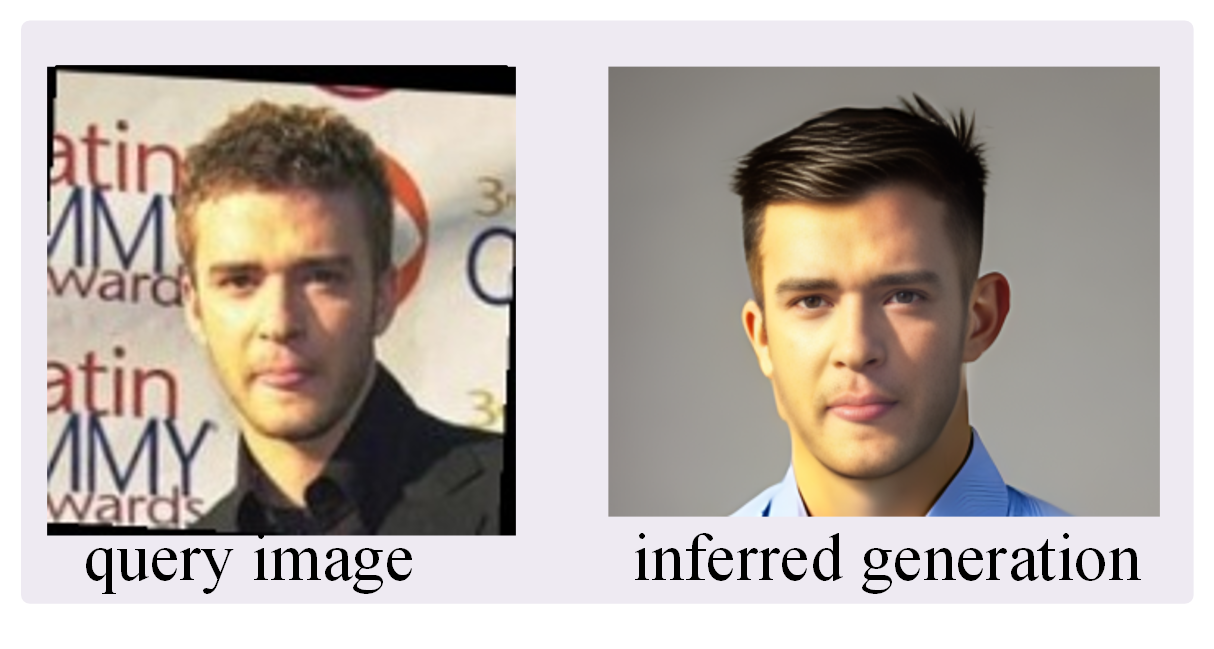}
    \caption{One example of data extraction success, where an image of Justin Timberlake was queried from the LFW dataset.}
    \label{fig:res_gen}
\end{figure}

 We explore the impact of sampling timesteps on the attack success rate across the three diffusion models, as described in Figure~\ref{fig:attack_success_rate}. 
 For this evaluation, we train new models on a merged dataset of CelebA and LFW, both resized to 128x128. 
 We vary the number of sampling timesteps for each diffusion model (DDPM, DDIM, LDM) and compare the attack success rate on different datasets. The attack success rate peaks at mid-range timesteps (100-150 for DDIM and LDM, 500-800 for DDPM). 
 However, for the LDM model, we observe the highest attack success rates across a broader range of timesteps.
 The results have been compiled in the  Table \ref{tab:iia}
 
\begin{table}[]
\resizebox{\textwidth}{!}{%

\centering
\begin{tabular}{cccccccc}
\hline
\multirow{2}{*}{\textbf{Diffusion Model}} & \multirow{2}{*}{\textbf{Sampling Steps}} & \multicolumn{3}{c|}{\textbf{CelebA}} & \multicolumn{3}{c}{\textbf{LFW}} \\ \cline{3-8} 
                         &   & \textbf{Accuracy} & \textbf{AUC} & \multicolumn{1}{c|}{} & \textbf{Accuracy} & \textbf{AUC} &  \\ \hline
DDPM                     & 1000                   & 0.76              & 0.72         & \multicolumn{1}{c|}{} & 0.85              & 0.80         &  \\ \hline
DDIM                     & 200                    & 0.73              & 0.68         & \multicolumn{1}{c|}{} & 0.82              & 0.77         &  \\ \hline
LDM                      & 200                    & 0.78              & 0.74         & \multicolumn{1}{c|}{} & 0.92              & 0.90         &  \\ \hline
\end{tabular}%
}
\caption{
Comparison of inference attack accuracy and AUC ROC across CelebA and LFW datasets for three diffusion models (DDPM, DDIM, LDM) at specific sampling steps (1000 for DDPM, 200 each for DDIM and LDM) for lists of 8 query images, over 7 experiment runs.}
\label{tab:iia}
\end{table}

\subsection{Data Extraction Attack}

In this subsection, we assess the efficacy of the data extraction attack by evaluating the Attack Success Rate (ASR) using two distinct metrics: ASR-one and ASR-MIA. ASR quantifies the proportion of queried images that successfully result in correct extractions. ASR-one pertains to cases where, among the top 10 identified candidate images, at least one is visually correlated with a data point in the training set. ASR-MIA integrates our membership inference attack (MIA) method with the data extraction attack (DEA), indicating that, for the same set of 10 representative images, at least one image was inferred as a member with a confidence score of 0.6 or higher.

Figure \ref{fig:res_gen} provides an illustrative example of a query image on the diffusion model trained on the LFW dataset alongside a generated image inferred as a match in the training dataset. 
\begin{table}[]

\centering
\begin{tabular}{cccc}
\hline
\textbf{Model} & \textbf{Timesteps} & \textbf{ASR-one} & \textbf{ASR-MIA} \\ \hline
\multirow{3}{*}{DDPM}& 50  & 82.1\% & 77.3\% \\ \cline{2-4}
 & 100 & 85.4\% & 78.9\% \\ \cline{2-4}
 & 200 & \textbf{91.6\%} & \textbf{82.1\%} \\ \hline

\multirow{3}{*}{DDIM} & 100  & 79.3\% & 75.2\% \\\cline{2-4}
& 200 & \textbf{89.2\%} & \textbf{81.2\% } \\ \cline{2-4}
& 400 & 84.8\% & 77.6\%\\ \hline

\multirow{3}{*}{LDM} & 100  & 76.4\% & 72.5\% \\ \cline{2-4}
& 200 & \textbf{87.5\%} & \textbf{79.5\% } \\ \cline{2-4} 
& 400 & 81.7\% & 76.1\% \\ \hline
\end{tabular}
\caption{Performance Evaluation of the Data Extraction Attack, reported on the LFW dataset.}
\label{tab:eval_results}
\end{table}
We further evaluated the performance of this attack across three different diffusion models—DDPM, DDIM, and LDM—by varying the number of sampling timesteps. Table \ref{tab:eval_results} presents these models' ASR-one and ASR-MIA results, along with the corresponding accuracy and AUC-ROC values. 

Evaluations as described in Table \ref{tab:eval_results} demonstrate that the success of the data extraction attack is highly dependent on the number of sampling timesteps. For the DDPM model, the highest ASR-one (91.6\%) and ASR-MIA (92.1\%) values were observed at 200 timesteps. In contrast, DDIM showed lower overall performance, achieving its best results at 200 timesteps with an ASR-one of 89.2\% 

These findings suggest that longer sampling sequences generally improve the attack’s effectiveness. Specifically, the DDPM model outperformed the others at higher timesteps, while DDIM lagged in ASR and AUC. LDM, while performing slightly worse than DDPM, provided a more consistent attack success rate across varying timesteps.

\section{Conclusions}
This paper introduced the novel concept of identity inference in trained diffusion models, providing a framework for holding model developers accountable for using personal data in training. Our identity inference attack moves beyond conventional membership inference, targeting the presence of an individual's identity in the training data, which offers a broader and more effective means of ensuring ethical data use. Additionally, our approach supports both single-query and multi-query scenarios, providing flexibility in its application. We also proposed a membership inference attack that surpasses current state-of-the-art techniques, offering superior accuracy in determining whether specific images were used in training.
\\
Looking ahead, a key direction for future research involves relaxing the white-box assumptions of our current framework. Transitioning to black-box settings will enhance the applicability of identity inference attacks, further strengthening the mechanisms for transparency and accountability in the deployment of diffusion models. This progression will bring us closer to robust, real-world tools for enforcing data privacy regulations in AI-driven systems.



\begin{thebibliography}{32}


\ifx \showCODEN    \undefined \def \showCODEN     #1{\unskip}     \fi
\ifx \showDOI      \undefined \def \showDOI       #1{#1}\fi
\ifx \showISBNx    \undefined \def \showISBNx     #1{\unskip}     \fi
\ifx \showISBNxiii \undefined \def \showISBNxiii  #1{\unskip}     \fi
\ifx \showISSN     \undefined \def \showISSN      #1{\unskip}     \fi
\ifx \showLCCN     \undefined \def \showLCCN      #1{\unskip}     \fi
\ifx \shownote     \undefined \def \shownote      #1{#1}          \fi
\ifx \showarticletitle \undefined \def \showarticletitle #1{#1}   \fi
\ifx \showURL      \undefined \def \showURL       {\relax}        \fi
\providecommand\bibfield[2]{#2}
\providecommand\bibinfo[2]{#2}
\providecommand\natexlab[1]{#1}
\providecommand\showeprint[2][]{arXiv:#2}

\bibitem[Bommasani et~al\mbox{.}(2021)]%
        {bommasani2021opportunities}
\bibfield{author}{\bibinfo{person}{Rishi Bommasani}, \bibinfo{person}{Drew~A Hudson}, \bibinfo{person}{Ehsan Adeli}, \bibinfo{person}{Russ Altman}, \bibinfo{person}{Simran Arora}, \bibinfo{person}{Sydney von Arx}, \bibinfo{person}{Michael~S Bernstein}, \bibinfo{person}{Jeannette Bohg}, \bibinfo{person}{Antoine Bosselut}, \bibinfo{person}{Emma Brunskill}, {et~al\mbox{.}}} \bibinfo{year}{2021}\natexlab{}.
\newblock \showarticletitle{On the opportunities and risks of foundation models}.
\newblock \bibinfo{journal}{\emph{arXiv preprint arXiv:2108.07258}} (\bibinfo{year}{2021}).
\newblock


\bibitem[Carlini et~al\mbox{.}(2023)]%
        {carlini2023extracting}
\bibfield{author}{\bibinfo{person}{Nicolas Carlini}, \bibinfo{person}{Jamie Hayes}, \bibinfo{person}{Milad Nasr}, \bibinfo{person}{Matthew Jagielski}, \bibinfo{person}{Vikash Sehwag}, \bibinfo{person}{Florian Tramer}, \bibinfo{person}{Borja Balle}, \bibinfo{person}{Daphne Ippolito}, {and} \bibinfo{person}{Eric Wallace}.} \bibinfo{year}{2023}\natexlab{}.
\newblock \showarticletitle{Extracting training data from diffusion models}. In \bibinfo{booktitle}{\emph{32nd USENIX Security Symposium (USENIX Security 23)}}. \bibinfo{pages}{5253--5270}.
\newblock


\bibitem[DeAlcala et~al\mbox{.}(2024)]%
        {dealcala2024my}
\bibfield{author}{\bibinfo{person}{Daniel DeAlcala}, \bibinfo{person}{Aythami Morales}, \bibinfo{person}{Gonzalo Mancera}, \bibinfo{person}{Julian Fierrez}, \bibinfo{person}{Ruben Tolosana}, {and} \bibinfo{person}{Javier Ortega-Garcia}.} \bibinfo{year}{2024}\natexlab{}.
\newblock \showarticletitle{Is my Data in your AI Model? Membership Inference Test with Application to Face Images}.
\newblock \bibinfo{journal}{\emph{arXiv preprint arXiv:2402.09225}} (\bibinfo{year}{2024}).
\newblock


\bibitem[Duan et~al\mbox{.}(2023)]%
        {duan2023diffusion}
\bibfield{author}{\bibinfo{person}{Jinhao Duan}, \bibinfo{person}{Fei Kong}, \bibinfo{person}{Shiqi Wang}, \bibinfo{person}{Xiaoshuang Shi}, {and} \bibinfo{person}{Kaidi Xu}.} \bibinfo{year}{2023}\natexlab{}.
\newblock \showarticletitle{Are diffusion models vulnerable to membership inference attacks?}. In \bibinfo{booktitle}{\emph{International Conference on Machine Learning}}. PMLR, \bibinfo{pages}{8717--8730}.
\newblock


\bibitem[Fu et~al\mbox{.}(2024)]%
        {fu2024model}
\bibfield{author}{\bibinfo{person}{Xiaomeng Fu}, \bibinfo{person}{Xi Wang}, \bibinfo{person}{Qiao Li}, \bibinfo{person}{Jin Liu}, \bibinfo{person}{Jiao Dai}, {and} \bibinfo{person}{Jizhong Han}.} \bibinfo{year}{2024}\natexlab{}.
\newblock \showarticletitle{Model Will Tell: Training Membership Inference for Diffusion Models}.
\newblock \bibinfo{journal}{\emph{arXiv preprint arXiv:2403.08487}} (\bibinfo{year}{2024}).
\newblock


\bibitem[Goodfellow et~al\mbox{.}(2020)]%
        {goodfellow2020generative}
\bibfield{author}{\bibinfo{person}{Ian Goodfellow}, \bibinfo{person}{Jean Pouget-Abadie}, \bibinfo{person}{Mehdi Mirza}, \bibinfo{person}{Bing Xu}, \bibinfo{person}{David Warde-Farley}, \bibinfo{person}{Sherjil Ozair}, \bibinfo{person}{Aaron Courville}, {and} \bibinfo{person}{Yoshua Bengio}.} \bibinfo{year}{2020}\natexlab{}.
\newblock \showarticletitle{Generative adversarial networks}.
\newblock \bibinfo{journal}{\emph{Commun. ACM}} \bibinfo{volume}{63}, \bibinfo{number}{11} (\bibinfo{year}{2020}), \bibinfo{pages}{139--144}.
\newblock


\bibitem[Ho et~al\mbox{.}(2020)]%
        {ho2020denoising}
\bibfield{author}{\bibinfo{person}{Jonathan Ho}, \bibinfo{person}{Ajay Jain}, {and} \bibinfo{person}{Pieter Abbeel}.} \bibinfo{year}{2020}\natexlab{}.
\newblock \showarticletitle{Denoising diffusion probabilistic models}.
\newblock \bibinfo{journal}{\emph{Advances in neural information processing systems}}  \bibinfo{volume}{33} (\bibinfo{year}{2020}), \bibinfo{pages}{6840--6851}.
\newblock


\bibitem[Ho et~al\mbox{.}(2022a)]%
        {ho2022cascaded}
\bibfield{author}{\bibinfo{person}{Jonathan Ho}, \bibinfo{person}{Chitwan Saharia}, \bibinfo{person}{William Chan}, \bibinfo{person}{David~J Fleet}, \bibinfo{person}{Mohammad Norouzi}, {and} \bibinfo{person}{Tim Salimans}.} \bibinfo{year}{2022}\natexlab{a}.
\newblock \showarticletitle{Cascaded diffusion models for high fidelity image generation}.
\newblock \bibinfo{journal}{\emph{Journal of Machine Learning Research}} \bibinfo{volume}{23}, \bibinfo{number}{47} (\bibinfo{year}{2022}), \bibinfo{pages}{1--33}.
\newblock


\bibitem[Ho et~al\mbox{.}(2022b)]%
        {ho2022video}
\bibfield{author}{\bibinfo{person}{Jonathan Ho}, \bibinfo{person}{Tim Salimans}, \bibinfo{person}{Alexey Gritsenko}, \bibinfo{person}{William Chan}, \bibinfo{person}{Mohammad Norouzi}, {and} \bibinfo{person}{David~J Fleet}.} \bibinfo{year}{2022}\natexlab{b}.
\newblock \showarticletitle{Video diffusion models}.
\newblock \bibinfo{journal}{\emph{Advances in Neural Information Processing Systems}}  \bibinfo{volume}{35} (\bibinfo{year}{2022}), \bibinfo{pages}{8633--8646}.
\newblock


\bibitem[Huang et~al\mbox{.}(2008)]%
        {huang2008labeled}
\bibfield{author}{\bibinfo{person}{Gary~B Huang}, \bibinfo{person}{Marwan Mattar}, \bibinfo{person}{Tamara Berg}, {and} \bibinfo{person}{Eric Learned-Miller}.} \bibinfo{year}{2008}\natexlab{}.
\newblock \showarticletitle{Labeled faces in the wild: A database forstudying face recognition in unconstrained environments}. In \bibinfo{booktitle}{\emph{Workshop on faces in'Real-Life'Images: detection, alignment, and recognition}}.
\newblock


\bibitem[Huang et~al\mbox{.}(2023)]%
        {huang2023collaborative}
\bibfield{author}{\bibinfo{person}{Ziqi Huang}, \bibinfo{person}{Kelvin~CK Chan}, \bibinfo{person}{Yuming Jiang}, {and} \bibinfo{person}{Ziwei Liu}.} \bibinfo{year}{2023}\natexlab{}.
\newblock \showarticletitle{Collaborative diffusion for multi-modal face generation and editing}. In \bibinfo{booktitle}{\emph{Proceedings of the IEEE/CVF Conference on Computer Vision and Pattern Recognition}}. \bibinfo{pages}{6080--6090}.
\newblock


\bibitem[Kim et~al\mbox{.}(2023)]%
        {kim2023dcface}
\bibfield{author}{\bibinfo{person}{Minchul Kim}, \bibinfo{person}{Feng Liu}, \bibinfo{person}{Anil Jain}, {and} \bibinfo{person}{Xiaoming Liu}.} \bibinfo{year}{2023}\natexlab{}.
\newblock \showarticletitle{Dcface: Synthetic face generation with dual condition diffusion model}. In \bibinfo{booktitle}{\emph{Proceedings of the ieee/cvf conference on computer vision and pattern recognition}}. \bibinfo{pages}{12715--12725}.
\newblock


\bibitem[Kingma(2013)]%
        {kingma2013auto}
\bibfield{author}{\bibinfo{person}{Diederik~P Kingma}.} \bibinfo{year}{2013}\natexlab{}.
\newblock \showarticletitle{Auto-encoding variational bayes}.
\newblock \bibinfo{journal}{\emph{arXiv preprint arXiv:1312.6114}} (\bibinfo{year}{2013}).
\newblock


\bibitem[Kong et~al\mbox{.}(2023)]%
        {kong2023efficient}
\bibfield{author}{\bibinfo{person}{Fei Kong}, \bibinfo{person}{Jinhao Duan}, \bibinfo{person}{RuiPeng Ma}, \bibinfo{person}{Hengtao Shen}, \bibinfo{person}{Xiaofeng Zhu}, \bibinfo{person}{Xiaoshuang Shi}, {and} \bibinfo{person}{Kaidi Xu}.} \bibinfo{year}{2023}\natexlab{}.
\newblock \showarticletitle{An efficient membership inference attack for the diffusion model by proximal initialization}.
\newblock \bibinfo{journal}{\emph{arXiv preprint arXiv:2305.18355}} (\bibinfo{year}{2023}).
\newblock


\bibitem[Li et~al\mbox{.}(2024a)]%
        {li2024towards}
\bibfield{author}{\bibinfo{person}{Jingwei Li}, \bibinfo{person}{Jing Dong}, \bibinfo{person}{Tianxing He}, {and} \bibinfo{person}{Jingzhao Zhang}.} \bibinfo{year}{2024}\natexlab{a}.
\newblock \showarticletitle{Towards Black-Box Membership Inference Attack for Diffusion Models}.
\newblock \bibinfo{journal}{\emph{arXiv preprint arXiv:2405.20771}} (\bibinfo{year}{2024}).
\newblock


\bibitem[Li et~al\mbox{.}(2024b)]%
        {li2024unveiling}
\bibfield{author}{\bibinfo{person}{Qiao Li}, \bibinfo{person}{Xiaomeng Fu}, \bibinfo{person}{Xi Wang}, \bibinfo{person}{Jin Liu}, \bibinfo{person}{Xingyu Gao}, \bibinfo{person}{Jiao Dai}, {and} \bibinfo{person}{Jizhong Han}.} \bibinfo{year}{2024}\natexlab{b}.
\newblock \showarticletitle{Unveiling Structural Memorization: Structural Membership Inference Attack for Text-to-Image Diffusion Models}.
\newblock \bibinfo{journal}{\emph{arXiv preprint arXiv:2407.13252}} (\bibinfo{year}{2024}).
\newblock


\bibitem[Liu et~al\mbox{.}(2024)]%
        {liu2024audioldm}
\bibfield{author}{\bibinfo{person}{Haohe Liu}, \bibinfo{person}{Yi Yuan}, \bibinfo{person}{Xubo Liu}, \bibinfo{person}{Xinhao Mei}, \bibinfo{person}{Qiuqiang Kong}, \bibinfo{person}{Qiao Tian}, \bibinfo{person}{Yuping Wang}, \bibinfo{person}{Wenwu Wang}, \bibinfo{person}{Yuxuan Wang}, {and} \bibinfo{person}{Mark~D Plumbley}.} \bibinfo{year}{2024}\natexlab{}.
\newblock \showarticletitle{Audioldm 2: Learning holistic audio generation with self-supervised pretraining}.
\newblock \bibinfo{journal}{\emph{IEEE/ACM Transactions on Audio, Speech, and Language Processing}} (\bibinfo{year}{2024}).
\newblock


\bibitem[Liu et~al\mbox{.}(2015)]%
        {liu2015deep}
\bibfield{author}{\bibinfo{person}{Ziwei Liu}, \bibinfo{person}{Ping Luo}, \bibinfo{person}{Xiaogang Wang}, {and} \bibinfo{person}{Xiaoou Tang}.} \bibinfo{year}{2015}\natexlab{}.
\newblock \showarticletitle{Deep learning face attributes in the wild}. In \bibinfo{booktitle}{\emph{Proceedings of the IEEE international conference on computer vision}}. \bibinfo{pages}{3730--3738}.
\newblock


\bibitem[Matsumoto et~al\mbox{.}(2023)]%
        {matsumoto2023membership}
\bibfield{author}{\bibinfo{person}{Tomoya Matsumoto}, \bibinfo{person}{Takayuki Miura}, {and} \bibinfo{person}{Naoto Yanai}.} \bibinfo{year}{2023}\natexlab{}.
\newblock \showarticletitle{Membership inference attacks against diffusion models}. In \bibinfo{booktitle}{\emph{2023 IEEE Security and Privacy Workshops (SPW)}}. IEEE, \bibinfo{pages}{77--83}.
\newblock


\bibitem[Miernicki and Ng(2021)]%
        {miernicki2021artificial}
\bibfield{author}{\bibinfo{person}{Martin Miernicki} {and} \bibinfo{person}{Irene Ng}.} \bibinfo{year}{2021}\natexlab{}.
\newblock \showarticletitle{Artificial intelligence and moral rights}.
\newblock \bibinfo{journal}{\emph{Ai \& Society}} \bibinfo{volume}{36}, \bibinfo{number}{1} (\bibinfo{year}{2021}), \bibinfo{pages}{319--329}.
\newblock


\bibitem[Pang and Wang(2023)]%
        {pang2023black}
\bibfield{author}{\bibinfo{person}{Yan Pang} {and} \bibinfo{person}{Tianhao Wang}.} \bibinfo{year}{2023}\natexlab{}.
\newblock \showarticletitle{Black-box membership inference attacks against fine-tuned diffusion models}.
\newblock \bibinfo{journal}{\emph{arXiv preprint arXiv:2312.08207}} (\bibinfo{year}{2023}).
\newblock


\bibitem[Pang et~al\mbox{.}(2023)]%
        {pang2023white}
\bibfield{author}{\bibinfo{person}{Yan Pang}, \bibinfo{person}{Tianhao Wang}, \bibinfo{person}{Xuhui Kang}, \bibinfo{person}{Mengdi Huai}, {and} \bibinfo{person}{Yang Zhang}.} \bibinfo{year}{2023}\natexlab{}.
\newblock \showarticletitle{White-box membership inference attacks against diffusion models}.
\newblock \bibinfo{journal}{\emph{arXiv preprint arXiv:2308.06405}} (\bibinfo{year}{2023}).
\newblock


\bibitem[Rombach et~al\mbox{.}(2022)]%
        {rombach2022high}
\bibfield{author}{\bibinfo{person}{Robin Rombach}, \bibinfo{person}{Andreas Blattmann}, \bibinfo{person}{Dominik Lorenz}, \bibinfo{person}{Patrick Esser}, {and} \bibinfo{person}{Bj{\"o}rn Ommer}.} \bibinfo{year}{2022}\natexlab{}.
\newblock \showarticletitle{High-resolution image synthesis with latent diffusion models}. In \bibinfo{booktitle}{\emph{Proceedings of the IEEE/CVF conference on computer vision and pattern recognition}}. \bibinfo{pages}{10684--10695}.
\newblock


\bibitem[Saharia et~al\mbox{.}(2022)]%
        {saharia2022palette}
\bibfield{author}{\bibinfo{person}{Chitwan Saharia}, \bibinfo{person}{William Chan}, \bibinfo{person}{Huiwen Chang}, \bibinfo{person}{Chris Lee}, \bibinfo{person}{Jonathan Ho}, \bibinfo{person}{Tim Salimans}, \bibinfo{person}{David Fleet}, {and} \bibinfo{person}{Mohammad Norouzi}.} \bibinfo{year}{2022}\natexlab{}.
\newblock \showarticletitle{Palette: Image-to-image diffusion models}. In \bibinfo{booktitle}{\emph{ACM SIGGRAPH 2022 conference proceedings}}. \bibinfo{pages}{1--10}.
\newblock


\bibitem[Shokri et~al\mbox{.}(2017)]%
        {shokri2017membership}
\bibfield{author}{\bibinfo{person}{Reza Shokri}, \bibinfo{person}{Marco Stronati}, \bibinfo{person}{Congzheng Song}, {and} \bibinfo{person}{Vitaly Shmatikov}.} \bibinfo{year}{2017}\natexlab{}.
\newblock \showarticletitle{Membership inference attacks against machine learning models}. In \bibinfo{booktitle}{\emph{2017 IEEE symposium on security and privacy (SP)}}. IEEE, \bibinfo{pages}{3--18}.
\newblock


\bibitem[Sohl-Dickstein et~al\mbox{.}(2015)]%
        {sohl2015deep}
\bibfield{author}{\bibinfo{person}{Jascha Sohl-Dickstein}, \bibinfo{person}{Eric Weiss}, \bibinfo{person}{Niru Maheswaranathan}, {and} \bibinfo{person}{Surya Ganguli}.} \bibinfo{year}{2015}\natexlab{}.
\newblock \showarticletitle{Deep unsupervised learning using nonequilibrium thermodynamics}. In \bibinfo{booktitle}{\emph{International conference on machine learning}}. PMLR, \bibinfo{pages}{2256--2265}.
\newblock


\bibitem[Song et~al\mbox{.}(2020)]%
        {song2020denoising}
\bibfield{author}{\bibinfo{person}{Jiaming Song}, \bibinfo{person}{Chenlin Meng}, {and} \bibinfo{person}{Stefano Ermon}.} \bibinfo{year}{2020}\natexlab{}.
\newblock \showarticletitle{Denoising diffusion implicit models}.
\newblock \bibinfo{journal}{\emph{arXiv preprint arXiv:2010.02502}} (\bibinfo{year}{2020}).
\newblock


\bibitem[Tang et~al\mbox{.}(2023)]%
        {tang2023membership}
\bibfield{author}{\bibinfo{person}{Shuai Tang}, \bibinfo{person}{Zhiwei~Steven Wu}, \bibinfo{person}{Sergul Aydore}, \bibinfo{person}{Michael Kearns}, {and} \bibinfo{person}{Aaron Roth}.} \bibinfo{year}{2023}\natexlab{}.
\newblock \showarticletitle{Membership inference attacks on diffusion models via quantile regression}.
\newblock \bibinfo{journal}{\emph{arXiv preprint arXiv:2312.05140}} (\bibinfo{year}{2023}).
\newblock


\bibitem[Wan et~al\mbox{.}(2023)]%
        {wan2023precise}
\bibfield{author}{\bibinfo{person}{Jun Wan}, \bibinfo{person}{Jun Liu}, \bibinfo{person}{Jie Zhou}, \bibinfo{person}{Zhihui Lai}, \bibinfo{person}{Linlin Shen}, \bibinfo{person}{Hang Sun}, \bibinfo{person}{Ping Xiong}, {and} \bibinfo{person}{Wenwen Min}.} \bibinfo{year}{2023}\natexlab{}.
\newblock \showarticletitle{Precise facial landmark detection by reference heatmap transformer}.
\newblock \bibinfo{journal}{\emph{IEEE Transactions on Image Processing}}  \bibinfo{volume}{32} (\bibinfo{year}{2023}), \bibinfo{pages}{1966--1977}.
\newblock


\bibitem[Wolleb et~al\mbox{.}(2022)]%
        {wolleb2022diffusion}
\bibfield{author}{\bibinfo{person}{Julia Wolleb}, \bibinfo{person}{Florentin Bieder}, \bibinfo{person}{Robin Sandk{\"u}hler}, {and} \bibinfo{person}{Philippe~C Cattin}.} \bibinfo{year}{2022}\natexlab{}.
\newblock \showarticletitle{Diffusion models for medical anomaly detection}. In \bibinfo{booktitle}{\emph{International Conference on Medical image computing and computer-assisted intervention}}. Springer, \bibinfo{pages}{35--45}.
\newblock


\bibitem[Wu et~al\mbox{.}(2022)]%
        {wu2022membership}
\bibfield{author}{\bibinfo{person}{Yixin Wu}, \bibinfo{person}{Ning Yu}, \bibinfo{person}{Zheng Li}, \bibinfo{person}{Michael Backes}, {and} \bibinfo{person}{Yang Zhang}.} \bibinfo{year}{2022}\natexlab{}.
\newblock \showarticletitle{Membership inference attacks against text-to-image generation models}.
\newblock  (\bibinfo{year}{2022}).
\newblock


\bibitem[Zhai et~al\mbox{.}(2024)]%
        {zhai2024membership}
\bibfield{author}{\bibinfo{person}{Shengfang Zhai}, \bibinfo{person}{Huanran Chen}, \bibinfo{person}{Yinpeng Dong}, \bibinfo{person}{Jiajun Li}, \bibinfo{person}{Qingni Shen}, \bibinfo{person}{Yansong Gao}, \bibinfo{person}{Hang Su}, {and} \bibinfo{person}{Yang Liu}.} \bibinfo{year}{2024}\natexlab{}.
\newblock \showarticletitle{Membership Inference on Text-to-Image Diffusion Models via Conditional Likelihood Discrepancy}.
\newblock \bibinfo{journal}{\emph{arXiv preprint arXiv:2405.14800}} (\bibinfo{year}{2024}).
\newblock


\end{thebibliography}
\end{document}